# RL-CoSeg：A Novel Image Co-Segmentation Algorithm with Deep Reinforcement Learning

Xin Duan[1], Xiabi Liu*, Xiaopeng Gong, Mengqiao Han

Beijing Lab of Intelligent Information Technology, School of Computer Science, Beijing Institute of Technology, Beijing 100081, China

Abstract

　　This paper proposes an automatic image co-segmentation algorithm based on deep reinforcement learning (RL). Existing co-segmentation tasks mainly rely on deep learning methods, and the obtained foreground edges are often rough. In order to obtain more precise foreground edges, we use deep RL to solve this problem and achieve the finer segmentation. To our best knowledge, this is the first work to apply RL methods to co-segmentation. We define the problem as a Markov Decision Process (MDP) and optimize it by RL with asynchronous advantage actor-critic (A3C). The RL image co-segmentation network uses the correlation between images to segment common and salient objects from a set of related images. In order to achieve automatic segmentation, our RL-CoSeg method eliminates user's hints. For the image co-segmentation problem, we propose a collaborative RL algorithm based on the A3C model. We propose a Siamese RL co-segmentation network structure to obtain the co-attention of images for co-segmentation. We improve the self-attention for automatic RL algorithm to obtain long-distance dependence and enlarge the receptive field. The image feature information obtained by self-attention can be used to supplement the deleted user's hints and help to obtain more accurate actions. Experimental results have shown that our method can improve the performance effectively on both coarse and fine initial segmentations, and it achieves the state-of-the-art performance on Internet dataset, iCoseg dataset and MLMR-COS dataset.

---

[1] E-mail:duanxin@bit.edu.cn

* Corresponding author.

1. Introduction

Segmenting the object of interest in an image is one of the fundamental problems in computer vision. Image co-segmentation is to segment common and salient objects from a set of related images. Since the concept was first proposed in 2006[1], it has attracted a lot of attention and many co-segmentation algorithms have been proposed. There are two main reasons for its importance. On the technique aspect[2], the correlation between images brings valuable cues for defining the interested objects and alleviates the ill-pose nature of segmentation. On the application aspect[2], image co-segmentation algorithms can be applied to and play crucial roles in various applications, such as internet image mining, image retrieval, video tracking, video segmentation[3][4], and etc.

Existing co-segmentation tasks are mainly based on deep learning methods. However, the contours in the objects segmented by using deep learning method are usually rough, for example, the edges of foreground objects are jagged, extra foreground pixels are extracted around the foreground edge, or real foreground pixels are lost near the foreground edge. Some researchers effort to improve the accuracy of foreground edges by improving the loss function. Zhen et al. [5] proposed a boundary-aware loss function to focus on pixels near the boundary. Liu et al. [6] used two weighted edge losses on the U-shaped segmentation network to preserve the spatial boundary information. Karimi and Salcudean [7] minimized the Hausdorff distance between two contours for image segmentation.

To get finer contours of foreground objects in segmentation, reinforcement learning methods attracted people's attention. Some segmentation methods using reinforcement learning (RL) have been proposed. These RL methods usually use user interaction to prompt segmentation[8], mostly in the form of points, scribbles and bounding boxes, and etc. The hints obtained from interaction are used as a new label to improve segmentation performance[8][9][10][11]. One of the key ideas of the interactive RL segmentation algorithm is to extract features of the object while meeting the user's intention. For objects with complex backgrounds, users often need to perform a lot of interaction as a hint of the segmentation process to improve the segmentation results obtained by the algorithm. Because one interaction might not ensure the segmentation accuracy. Therefore, many existing methods use an iteratively-refined model: the operator provides new hints according to the current result to refine the segmentation until it is satisfactory[8]. Therefore, it is particularly important to reduce manpower while obtaining accurate segmentation. Gulshan et al. [12] uses the workload of users to measure system performance. Moreover, to reduce the number of interactions, the existing works replace the initial hints with an automatically-obtained coarse segmentation[8][9][11]. Obviously, the interactive RL segmentation algorithm cannot achieve automatic segmentation, and it can't meet the needs of batch segmentation because of its low efficiency in practical applications.

To overcome this drawback, we propose an automatic deep RL co-segmentation algorithm based on iterative update strategy, called RL-CoSeg for short. As far as we know, this is the first work that RL method is used to solve image co-segmentation problems. RL-CoSeg used iteratively-refined strategy. We use a Markov Decision Process (MDP) to represent this dynamic segmentation process. In each iteration, the model outputs actions to adjust the segmentation probability at each pixel, according to the pair of segmentation probabilities generated in the previous iteration. These actions represent different adjustment values, which are used to adjust the previous segmentation probability, so that the segmentation probability is closer to the label, and the segmentation will be more precise and finer [8]. In this process, the pair of original images are input into the co-attention block to obtain the co-attention vectors of the salient objects in the two images. The vectors are multiplied by the feature map of the segmentation probability to highlight the common salient objects of the two images. Inspired by non-local attention, we designed an attention block. The original image is input into it to get the attention matrix of the original image itself. The attention matrix is applied to the feature map of the segmentation probability to make the network focus on the feature and location of the salient object in the segmentation probability. Because the original image and its corresponding segmentation probability have the same object. We input the obtained new segmentation probability as feedback information to the model, and iterate this process repeatedly until a better segmentation result is obtained. Our RL-CoSeg method can effectively improve the performance of segmentation edges.

The experimental results show that RL-CoSeg method can improve the segmentation performance regardless of the coarse or fine initial segmentation. Even if the current best segmentation is used as the initial input, our proposed co-segmentation algorithm can still improve the segmentation performance. As far as we know, it achieves the state-of-the-art performance on Internet dataset and iCoseg dataset. We summarize our contributions as follows:

1. We use reinforcement learning methods to optimize co-segmentation tasks. Based on the A3C model, we propose a collaborative reinforcement learning algorithm.

2. We propose a deep RL co-segmentation Siamese network structure that jointly processes segmentation probability maps and original images. Each branch of the network structure has two output heads to output actions and value respectively.

3. We improve the self-attention for automatic RL image co-segmentation method. The image feature obtained by self-attention can be used to supplement the deleted user's hints in our RL-CoSeg method and capture long-range dependencies to obtain more accurate actions.

2. Related Work

2.1 Image co-segmentation algorithms

Since the emergence of the concept of co-segmentation, many traditional co-segmentation methods have been proposed. Optimize the objective function which is to describe the relationship between images to determine the label of each element in each image is a classic method in traditional co-segmentation[13][14][15][16][17]. Rubio et al. [18] and Kim et al. [19] suggested co-segmentation methods which employ region correspondence. Jerripothula et al. [21] and Rother et al. [22] tried to find out the optimal models which are specific to single images and vary across images to represent object regions or contours. Rubinstein et al.[20] proposed to combine visual saliency and dense pixel correspondences across images for co-segmentation.

Recently, we witness the applications of deep neural networks to image co-segmentation. Han et al. [24] constructed two graphs based on low-level visual features and high semantic features extracted from CNN. Wang et al. [25] used fully convolutional network to obtain the initial segmentation result on each single image. Yuan et al. [26] used a deep network to describe the dense conditional random fields (DCRF) for common objects. This deep network was used to calculate the probability of each pixel in the foreground[27], which was used in the second DCRF procedure to determine the final label. Li et al. [28] applied the Siamese network to image co-segmentation, and trained the network end-to-end. They used the U-shape network structure which encodes features by down-sampling and then up-sampling to get results from deep features.

2.2 Deep reinforcement learning

Deep reinforcement learning (RL) has been widely researched and applied. Mnih et al. [29][30] combined the convolutional neural network with the traditional Q-learning algorithm [31] in RL, and proposed a Deep Q-Network (DQN) model, which is the pioneering work in deep RL. The satisfactory performance of deep RL through DQN in Atari game has attracted many researchers' attention[29][30]. Hasselt et al. [33] proposed double Q-learning, and on this basis, they proposed the Deep Double Q-Network (DDQN) [33] algorithm. The dueling DQN algorithm was proposed by Wang et al [34]. In addition to Q-learning methods, deep RL methods based on policy gradients are also commonly used. The policy gradients method updates the policy parameters by calculating the gradient of the policy parameters with respect to the expected total rewards of the policy, and finally converges to the optimal policy[35]. Silver et al. [36] proposed Deterministic Policy Gradient （DPG） method. Lillicrap et al. [37] used the idea of DQN to improve the Deterministic Policy Gradient (DPG) method and proposed the Deep Deterministic Policy Gradient (DDPG) algorithm based on the Actor-Critic (AC) framework. Mnih et al. [38] proposed Asynchronous Advantage Actor-Critic（A3C）algorithm. The A3C algorithm uses multi-core CPUs to execute multiple agents in parallel and asynchronously. This asynchronous method can reduce training time and can well replace the experience playback mechanism. Our model is based on the A3C algorithm.

2.3 Application of deep reinforcement learning in image segmentation

Recently, RL is used in some image or video segmentation tasks. Liao et al. [8] used a multi-step iterative and interactive RL algorithm for medical image segmentation. Ghajari et al. [39] used multi-agent dimensional structure to segment the ultrasound image. Song et al. [40] used an RL agent to simulate the user's hints to meet the needs of interactive segmentation with less manpower. Han et al. [41] applied RL to video object segmentation. Agents are learned to segment object regions under a deep RL framework. Vecchio et al. [42] collected user clicks and use RL agents instead of humans to perform unsupervised video target segmentation. Most of the RL segmentation algorithms use interactive segmentation methods, which need to provide human hints to the segmentation system. The interactive segmentation method isn't feasible in large-scale applications. As far as we know, we are the first to propose an automatic multi-step iterative RL segmentation algorithm without human hints.

## 3. The proposed Method

There is still room for improvement in the accuracy of the segmentation obtained by the deep learning-based co-segmentation algorithm. We always observe rough contours in the objects segmented by deep networks with deep learning method, for example, the extra foreground pixels are extracted around the foreground edges or the true foreground pixels near the foreground edges are lost. And the segmentation results usually lose some details compared with the original images. To solve the above problems, we use deep reinforcement learning (RL) method to optimize the initial rough segmentation results.

In this method, we will get more accurate segmentation results by iteratively optimizing the rough initial segmentation probability map [8]. The initial rough segmentation probability map can be obtained by any segmentation method. In this paper, we use the co-segmentation method based on deep learning proposed by Gong et al. [15] to obtain the initial segmentation probability map. In each iteration, the model concatenates the segmentation probability map generated in the previous iteration with the original map according to the channel dimension, and input them to the branch of the Siamese network to obtain the joint feature map. The Siamese network adjusts the segmentation probability of each pixel with actions according to the joint feature map. These actions represent different adjustment values. Thus, the segmentation is closer to the true binary label mask, and the iterative process is carried out for many times to obtain more accurate segmentation.

The method of adjusting the pixel value at each pixel position by the action of reinforcement learning network to achieve the expected effect of image processing was first proposed by Furuta et al[45] and is used in image processing fields such as image denoising; Liao et al[8] used reinforcement learning network to output adjustment action for each voxel position in three-dimensional medical image for medical image segmentation.

In our method, the deep reinforcement learning Siamese network first obtains a joint feature map by extracting features on the concatenated image of the initial segmentation map and the original image, and then generates actions according to the joint feature map to adjust the segmentation probability. To obtain the correlation of a pair of images, we use the correlation block to calculate the co-attention of them, which provides more extra information for RL method. Agents for two images will be affected by the correlated features of the two related image environments, which can be seen as a correlated A3C algorithm. For automatic image segmentation, we delete the user interaction which is usually used in RL segmentation method. To compensate for the lack of user interaction, we improve the self-attention module according to the needs of RL method to capture more image information, which is utilized to obtain the long-distance dependencies of each image. Self-attention is applied to the feature map of segmentation probability map to focus on the target region of the image, so as to get accurate actions and fine segmentation. And the usage of feature map can reduce the computing resources. Our RL-CoSeg method can obtain finer co-segmentation results that retain more image details, especially on the edge of segmentation.

**3.1 Correlated Reinforcement learning algorithm for co-segmentation**

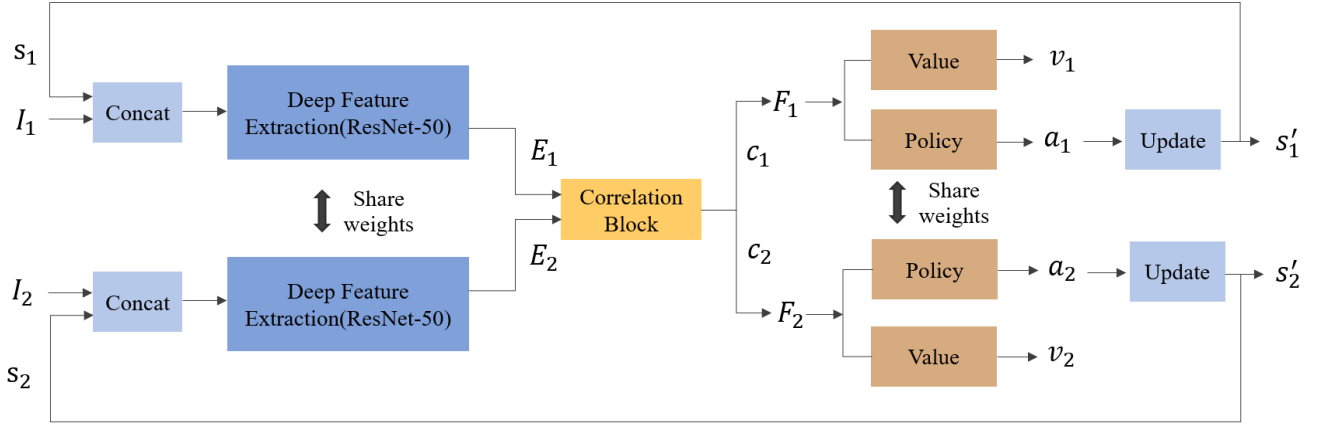

Figure 1. The collaborative computing process of the deep reinforcement learning co-segmentation method. Concatenate the original image with the segmentation probability map and obtain a joint feature map. Calculate the correlation of the joint feature map and obtaining a new segmentation probability map. (This section only introduces the collaborative reinforcement learning algorithm, so the self-attention calculation part is not introduced. In order to facilitate the understanding of the collaborative algorithm, the self-attention module is not represented in Figure 1.)

The traditional A3C reinforcement learning algorithm is aimed at the objects of the same environment. The action, state and reward of the agent are only affected by the current environment. We utilize the A3C algorithm to deal with the objects in the two related images at the same time for the co-segmentation problem, and each segmentation probability map is regarded as a RL environment. Due to the correlation of images in co-segmentation, the action, state and reward of agents for two segmentation probability maps will be affected by the correlated features of the two related image environments, so a correlated A3C reinforcement learning algorithm is obtained.

The process of deep RL co-segmentation algorithm is shown in Figure 1. In order to obtain the correlation of a pair of images, we need to calculate the co-attention between them. The 'co-attention' here means that we obtain the co-attention features in encoded features of an image to guide the attention in another image [43]. To obtain the co-attention map of two correlated input concatenated maps of segmentation probability maps and original images, we choose the Siamese encoder structure to map them to the same feature space, where the two encoder streams share weights. For policy and value networks, we also adopt the Siamese structure, in which two policy networks share weights, and two value networks share weights. We use ResNet50 to extract the deep features of the concatenated map of original image and the segmentation probability map to obtain the joint feature maps of them respectively, and then use the Correlation block[43] to calculate the co-attention of the joint feature maps.

We use the segmentation probability map as the state of the RL algorithm. And the policy network generates actions according to the joint feature map of the original image and segmentation probability map. These actions update the state to obtain a new one by fine-tuning the previous segmentation probability. The obtained new state is input to the deep feature extraction layer of the network again for a new iteration.

The co-attention map contains the attention of common object features of the two related images. Apply it to the two joint feature maps and the RL algorithm gets the new states of the two image environments. The new states are affected by the common features of both images, that means the new state of each image environment is affected by another environment. Therefore, the action and reward of each environment are also affected by another one. Thus our A3C algorithm for co-segmentation is correlated.

Specifically, for a pair of images $I_1$ and $I_2$, first, they are concatenated with the initial segmentation probability maps, which are the states of reinforcement learning $s_1$ and $s_2$. We use a correlation block to obtain the co-attention map of the encoded joint feature maps of them. $E_1$ and $E_2$ respectively represent the joint feature maps. We use $Y_1$ and $Y_2$ to represent the co-attention descriptors of $E_1$ and $E_2$ respectively, and $c_1$ and $c_2$ are the resultant correlation maps from co-attention. The overall procedure of this computation can be described formally as:

$$c_1 = E_1 \odot Y_2$$

$$c_2 = E_2 \odot Y_1$$

We apply $c_1$ and $c_2$ to the joint feature maps $E_1$ and $E_2$, and obtain the new joint feature maps $F_1$ and $F_2$ with the common features highlighted. The policy networks obtain their corresponding actions $a_1$ and $a_2$ to adjust the states according to the joint feature maps $F_1$ and $F_2$, respectively. $a_1$ and $a_2$ adjust the two states respectively to obtain new states $s'_1$ and $s'_2$ respectively. This process is shown in the fig 1.

Because the process of obtaining $a_1$ and $a_2$ is jointly influenced by states $s_1$ and $s_2$, and determined by the common features of $s_1$ and $s_2$. So we think that the probability of obtaining action $a_1$ is not only affected by the state $s_1$, but also related to the state $s_2$. Similarly, due to the correlation between images, we believe that the probability of obtaining $a_2$ is also affected by the both states. Therefore, in our RL co-segmentation model based on A3C, the policy of the RL agent for the images $I_1$ and $I_2$ are respectively:

$$\pi(a_1^t|s_1^t, s_2^t; \theta_1)$$

$$\pi(a_2^t|s_2^t, s_1^t; \theta_2)$$

For the value function, in the RL environment of the segmentation probability map for image $I_1$, due to the co-attention mechanism of $I_1$ and $I_2$, the value function of agents is affected by the state of image $I_2$ in training, so its value function is $V(s_1^t, s_2^t; \theta_{1v})$. Similarly, in the RL environment of image $I_2$, the value function of agents is $V(s_2^t, s_1^t; \theta_{2v})$. Therefore, in the A3C model, the gradient of the parameter vectors θ and $\theta_v$ of the policy function and value function for the images $I_1$ and $I_2$ is calculated as:

$$d\theta_1 \leftarrow d\theta_1 + \nabla_\theta \log\pi(a_1^t|s_1^t, s_2^t; \theta_1)(R - V(s_1^t, s_2^t; \theta_{1v}))$$

$$d\theta_{1v} \leftarrow d\theta_{1v} + \partial(R - V(s_1^t, s_2^t; \theta_{1v}))^2/\partial\theta_{1v}$$

$$d\theta_2 \leftarrow d\theta_2 + \nabla_\theta \log\pi(a_2^t|s_2^t, s_1^t; \theta_2)(R - V(s_2^t, s_1^t; \theta_{2v}))$$

$$d\theta_{2v} \leftarrow d\theta_{2v} + \partial(R - V(s_2^t, s_1^t; \theta_{2v}))^2/\partial\theta_{2v}$$

The pseudo code of the A3C model of reinforcement learning applied to co-segmentation is:

// Assume global shared parameter vectors $\theta_1$ and $\theta_{1v}$, $\theta_2$ and $\theta_{2v}$, global shared counter T = 0

// Assume thread-specific parameter vectors $\theta'_1$ and $\theta'_{1v}$, $\theta'_2$ and $\theta'_{2v}$

Initialize thread step counter $t \leftarrow 1$

repeat

    Reset gradients: $d\theta_1 \leftarrow 0$ and $d\theta_{1v} \leftarrow 0$, $d\theta_2 \leftarrow 0$ and $d\theta_{2v} \leftarrow 0$.

    Synchronize thread-specific parameters $\theta'_1 = \theta_1$, $\theta'_{1v} = \theta_{1v}$ and $\theta'_2 = \theta_2$, $\theta'_{2v} = \theta_{2v}$

    tstart = t

    Get state $s_1^t, s_2^t$

    repeat

        Perform $a_1^t$ according to policy $\pi(a_1^t | s_1^t, s_2^t; \theta'_1)$

        Receive reward $r_1^t$ and new state $s_1^{t+1}$

        Perform $a_2^t$ according to policy $\pi(a_2^t | s_2^t, s_1^t; \theta'_2)$

        Receive reward $r_2^t$ and new state $s_2^{t+1}$

        $t \leftarrow t + 1$

        $T \leftarrow T + 1$

    **until** terminal $s_1^{t+1}$, $s_2^{t+1}$ or t - tstart == tmax

$$R_1 = \begin{cases} 0, & \text{for terminal } s_1^t \\ V(s_1^t, s_2^t; \theta'_{1v}), & \text{for non} - \text{terminal } s_1^t \text{ //Bootstrap from last state} \end{cases}$$

$$R_2 = \begin{cases} 0, & \text{for terminal } s_2^t \\ V(s_2^t, s_1^t; \theta'_{2v}), & \text{for non} - \text{terminal } s_2^t \text{ //Bootstrap from last state} \end{cases}$$

    **for** $i \in \{t-1, \dots, t_{start}\}$ **do**

      $R_1 \leftarrow r_1^i + \gamma R_1$

      $R_2 \leftarrow r_2^i + \gamma R_2$

        Accumulate gradients wrt $\theta'_1$: $d\theta_1 \leftarrow d\theta_1 + \nabla_{\theta'_1} \log \pi(a_1^i | s_1^i, s_2^i; \theta'_1)(R_1 - V(s_1^i, s_2^i; \theta'_{1v}))$

        Accumulate gradients wrt $\theta'_{1v}$: $d\theta_{1v} \leftarrow d\theta_{1v} + \partial(R_1 - V(s_1^i, s_2^i; \theta'_{1v}))^2 / \partial \theta'_{1v}$

        Accumulate gradients wrt $\theta'_2$: $d\theta_2 \leftarrow d\theta_2 + \nabla_{\theta'_2} \log \pi(a_2^i | s_2^i, s_1^i; \theta'_2)(R_2 - V(s_2^i, s_1^i; \theta'_{2v}))$

        Accumulate gradients wrt $\theta'_{2v}$: $d\theta_{2v} \leftarrow d\theta_{2v} + \partial(R_2 - V(s_2^i, s_1^i; \theta'_{2v}))^2 / \partial \theta'_{2v}$

    **end for**

    Perform asynchronous update of $\theta_1$ using $d\theta_1$ and of $\theta_{1v}$ using $d\theta_{1v}$

    Perform asynchronous update of $\theta_2$ using $d\theta_2$ and of $\theta_{2v}$ using $d\theta_{2v}$

**Until** T > Tmax

## 3.2 Deep Reinforcement Learning Co-Segmentation Siamese Network Structure

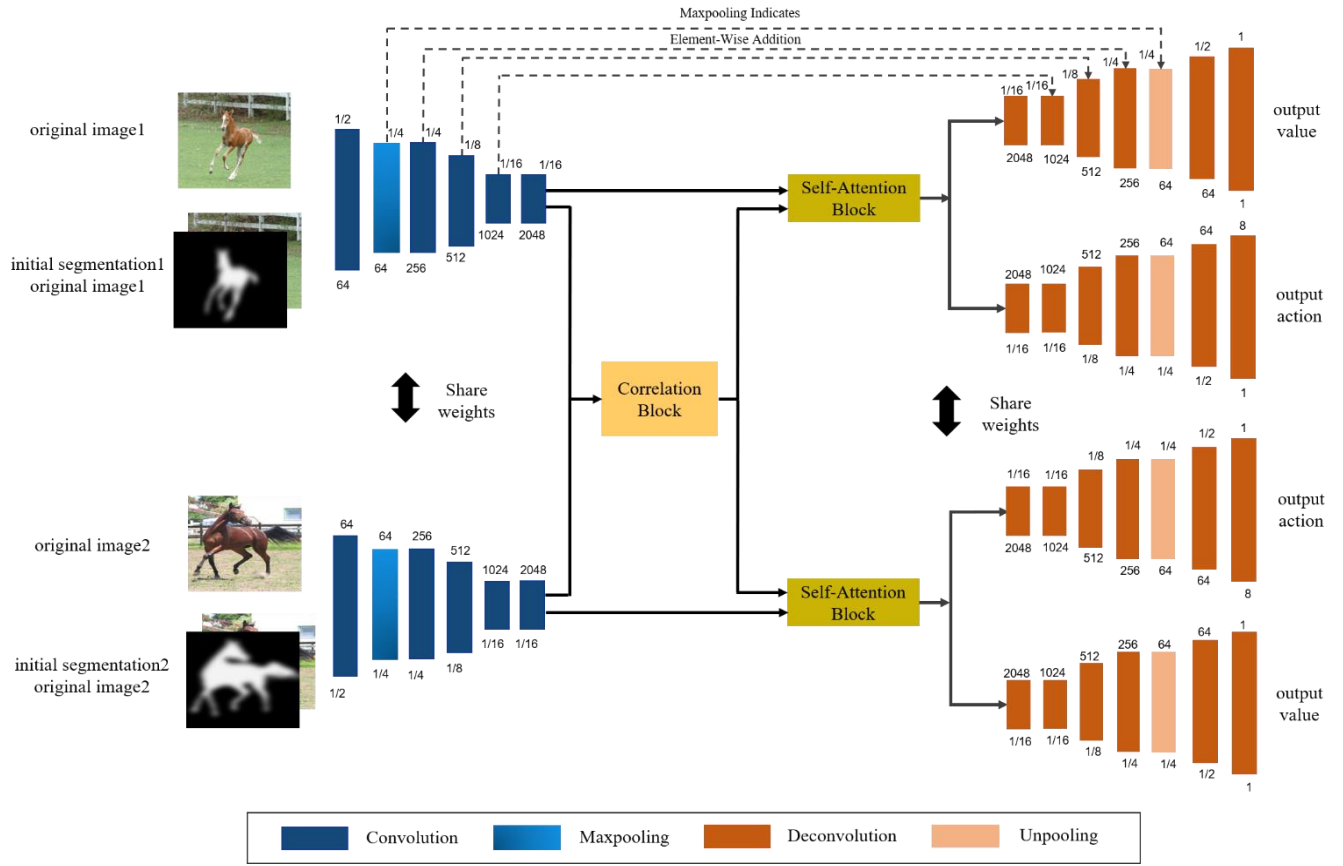

Figure 2. Network structure of deep reinforcement learning co-segmentation

The deep reinforcement learning Siamese network structure for image co-segmentation is shown in Figure 2. The deep reinforcement learning Siamese network we proposed includes a correlation calculation part and a self-attention calculation part. The input of the correlation calculation part is the concatenated initial segmentation probability map and the original image; the input of the self-attention calculation part is the original image. The correlation calculation part and the self-attention calculation part share the encoder of the two branches of the Siamese network, that is, the input of the correlation calculation part and the input of the self-attention calculation part use the encoder of the Siamese network for feature extraction in turn. And each encoder branch obtains two feature maps, which are the joint feature map of the concatenated segmentation probability map and the original image, as well as the feature map of the original image. The joint feature map is used for the calculation of the correlation module, and the feature map of the original image is used for the calculation of the self-attention module together with the joint feature map. We apply both the correlation map and the self-attention map to the joint feature map. The policy network and the value network of each branch of the Siamese network will generate actions and values according to their corresponding joint feature maps. We use ResNet-50 to build the Siamese encoder. In the following, we will explain the theory of the network structure.

For image co-segmentation, it is necessary to extract the co-attention of a pair of related images to obtain the correlation between them. In order to obtain the co-attention, we need to map the input to the same feature space, so we constructed a RL co-segmentation network with Siamese structure [44], as shown in Figure 2. It can obtain the image correlation for co-segmentation with a large receptive field, and it has the policy branch and value branch of the A3C RL algorithm. We use ResNet50 to extract deep features of the original image and the segmentation probability map.

We use a correlation block [43] to obtain the correlation of two concatenated inputs, and use the semantic joint feature maps to calculate a correlation map based on co-attention. In this way, the semantics of objects and the co-attention between objects are jointly utilized to detect common and salient objects across the images. In addition, we use the self-attention module to obtain the self-attention of the original image, aiming to use the self-attention to guide the segmentation probability feature map, so as to focus on the foreground target feature of the segmentation probability feature map and capture long-range dependencies.

For the self-attention calculation part, we calculate self-attention from the original image. Since our RL co-segmentation network automatically performs co-segmentation, for automatic segmentation, we remove the human interaction that is often used in RL segmentation tasks, but the hint map obtained by human interaction is used as an extra label in the segmentation process, which provides additional image information. Since there is no human interaction, in order to supplement more image feature information and highlight features of the objects contained in each image, besides co-attention, we add a self-attention module, which generates the self-attention map of the original image. The self-attention map is applied to the joint feature map.

To obtain the policy and value of the two images respectively, we input the segmentation probability map as the state into the network in the current iteration. And sample for actions according to the policy and value. Each pixel of the segmentation probability map corresponds to an action, which fine-tunes the segmentation probability at that pixel location [45]. We iterate and loop this process until we get a better segmentation result. In the first iteration, we use the initial segmentation probability as the initial input for the network.

### 3.3 Self-attention mechanism

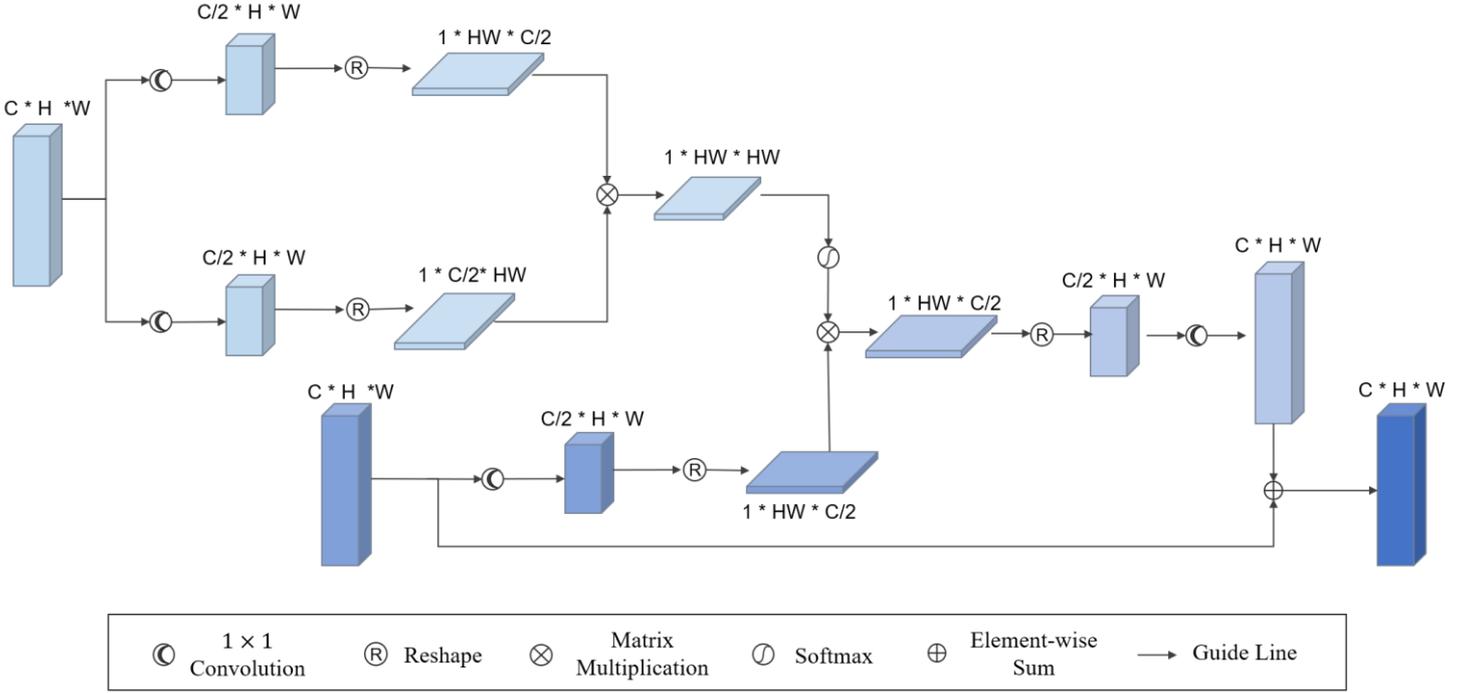

Figure 3. self-attention structure diagram.

Our network is an automatic segmentation network. For automatic segmentation, we remove the user interaction that is often used in RL segmentation tasks. User interaction is often used as an additional label for segmentation. Since there is no user interaction, it's necessary to add more image feature information and focus on the features of the objects in each image in order to get appropriate actions to adjust the segmentation probability. Moreover, one of the recent trends in CNNs is to enlarge the receptive field to boost the network performance. Based on the above two reasons, in addition to co-attention, we add a self-attention block to capture self-attention of each image.

Inspired by self-attention [48], the network captures the long-distance dependence of the original image and apply it to the feature map of the segmentation probability, so as to focus on the salient object on the segmentation probability feature map, and then get more appropriate actions and value to adjust the segmentation probability. The self-attention mechanism calculates the response between each pixel and all other pixels on the original image to express the similarity between pixels. Self-attention focus on the similar pixels in the image and ignore the pixels with low similarity. At the same time, it can capture the long-distance dependence of the image and expand the receptive field.

In our task, the network needs to obtain the action and value through the segmentation probability feature map to adjust the segmentation probability, so it needs to pay attention to the image feature area on the segmentation probability feature map. Since in the original image, the image features are more obvious, the details of the object are richer, the similarity and difference between pixels are more accurate, and the distribution of the image feature in the original image and the corresponding segmentation probability map is consistent, we use self-attention to obtain the long-distance dependency relationship of the original image and apply it to the feature map of the segmentation probability. In this way, the long-distance dependency relationship of the segmentation probability is highlighted by the original image, which is to supplement our removed user interaction with the image feature information of the original image. The self-attention structure for our RL-CoSeg method is shown in Figure 4. In this way, the network pays more attention to the object of the segmentation probability in the current state to obtain finer segmentation results.

Specifically, given a feature map $E$ of the original image and its corresponding joint feature map $P$, we calculate the self-attention of $E$, and multiply it by the joint feature map $P$, finally obtain the new joint feature map $P'$ which is highlighted by the self-attention. The calculation process is formulated as follows:

$$f(E) = w_1(e_n)^T w_2(e_m)$$

$$g(E) = w_3(p_n)$$

$$P' = w_4\big(\sigma(f(E))g(E)\big) + P$$

Among them, $w_1$, $w_2$, $w_3$, $w_4$ respectively represent the parameters of a 1*1 convolution kernel; $\sigma(\cdot)$ represents the activation function (softmax is used in this paper); $e_n$ is the output position whose response is to be computed in the feature map E, and $e_m$ represents all possible locations of E. f represents the obtained correlation between $e_n$ and $e_m$. $g(E)$ calculates the representation at position n on the joint feature map. $P'$ is the final highlighted joint feature map.

### 3.4 Fine-tune segmentation probabilities with reinforcement learning

Reinforcement learning (RL) has achieved satisfactory results in image denoising [45] and medical image segmentation [8]. We use RL methods to optimize the initial segmentation probability. Inspired by the application of RL methods for medical image segmentation [8], we apply it to the co-segmentation of natural images and construct an automatic segmentation model based on A3C algorithm. The policy branch of our RL co-segmentation network outputs actions, and the value network outputs value. The RL segmentation method is to obtain the adjusted value for the segmentation probability of each pixel by the agent, and then update the segmentation probability until get a satisfactory segmentation. This problem is defined as a Markov Decision Process (MDP) composed of state, action, and reward.

The agent executes the action according to the current state and receives the corresponding reward. Let $l_i$ be the i-th pixel of the input image $L$ with N pixels (i=1,...,N). Each pixel has a corresponding agent, defined by the strategy $\pi_i(a_i^{(t)}|s_i^{(t)})$, $\pi^{(t)} = \big(\pi_1^{(t)}, \ldots, \pi_N^{(t)}\big)$, $a_i^{(t)} \in A$, $s_i^{(t)} \in S$. $a_i^{(t)}$ and $s_i^{(t)}$ are the action and state of the i-th agent in the t-th step, A is the action set, and S is the state set. The action $a_i^{(t)}$ is the adjusted value of the segmentation probability of each $l_i$ (i=1,...,N), and the state $s_i^{(t)}$ is the previous segmentation probability of each $l_i$. The agent takes action $a_i^{(t)}$ to update the segmentation probability of pixel $l_i$ to obtain a new state $s_i^{(t)}$. The previous segmentation probability will be adjusted to a new one, and the reward $r_i^{(t)}$ is obtained from the environment to evaluate the segmentation result。 $p^{(t)}$ denotes the previous segmentation probability at time t, $p^{(t)} = (p_1^{(t)}, ..., p_N^{(t)})$ , $p_i^{(t)}$ is the segmentation probability of the i-th pixel at time t。 The action $a_i^{(t)}$ is discrete, it uses a certain probability value $d_i^{(t)}$ to adjust the previous segmentation probability $p_i^{(t)}$. After taking action $a_i^{(t)}$, the segmentation probability $p_i^{(t+1)}$ and state $s_i^{(t+1)}$ are：

$$p_i^{(t+1)} = F_0^1\left(p_i^{(t)} + d_i^{(t)} + c_i^{(t)}\right) \quad (1)$$

$$F_a^b(x) = min(max(x,a),b) \quad (2)$$

$$s_i^{(t+1)} = p_i^{(t+1)} \quad (3)$$

$F_a^b(x)$ is a function to clip the value of x from a to b. Since $p_i^{(t+1)}$ represents a probability, it is limited to [0,1]. The reward function is the relative improvement from previous segmentation probability to the current one，which is the difference between the probability and the squared error of groundtruth $yi$. Immediate rewards and accumulated rewards are described as follows：

$$r_i^{(t)} = (y_i - p_i^{(t-1)})^2 - (y_i - p_i^{(t)})^2 \quad (4)$$

$$R_i^{(t)} = r_i^{(t)} + \gamma r_i^{(t+1)} + \gamma^2 r_i^{(t+2)} + \cdots + \gamma^n r_i^{(t+n)} \quad (5)$$

Where $r_i^{(t)}$ is the immediate reward of a pixel at step t, $\gamma$ is the discount factor, and $R_i^{(t)}$ is the accumulated reward.

## 4. Experiments

We conduct three groups of experiments to evaluate the effectiveness of our proposed reinforcement learning (RL) co-segmentation network. We tested the performance of our proposed RL co-segmentation network on the iCoSeg dataset, Internet dataset, MLMR-COS dataset and compared it with other state-of-the-art techniques including Gong et al [43], Han et al. [24], Faktor and Jerripothula [46], Yuan et al [26] and Li et al [47]. On the experimental data sets used in this paper, the previous best performance among deep methods or traditional non-deep methods. Ablation experiments are used to prove the effectiveness of our RL co-segmentation algorithm.

### 4.1 Experimental setup

**4.1.1 datasets**

We use Pascal VOC 2012 and MSRC datasets to train our RL co-segmentation network, and then use Internet and iCoseg as the test set. These four data sets are widely used in the community of image co-segmentation. We further tested our network on the MLMR-COS dataset, which is suitable for co-segmentation. MLMR-COS dataset has simple backgrounds generated by single color. We use MLMR-COS to verify that our segmentation network can improve the segmentation results on the fine initial segmentation.

MSRC is composed of 591 images of 21 object groups. The ground-truth is roughly labeled, which does not align exactly with the object boundaries. VOC 2012 include 11540 images with ground-truth detection boxes and 2913 images with segmentation masks. Only 2913 images with segmentation masks can be considered in our problem. Note that not all of the examples in these two datasets can be used. In VOC 2012, the interested objects in some images have great changes in appearance and are cluttered in many other objects, so that the meaningful correlation between them is too hard to be found. The remaining 1743 images in VOC 2012 and 507 images in MSRC are used to construct our training set. From the training images, we sampled 13200 pairs of images containing common objects as our training set. We use the output of co-segmentation network proposed by Gong et al as the initial segmentation for our RL segmentation network.

iCoseg dataset contains 643 images divided into 38 object groups. Each group contains 17 images on average. The pixel-wise hand-annotated ground truth is offered. The backgrounds in each group are consistent natural scenes. Besides entire iCoseg dataset, many previous works use various versions of its subsets. Internet consists of 3 classes (airplane, car, and house) of thousands of downloaded Internet images. Following the compared methods, we evaluate our approach on its widely used subset, in which each class has 100 images.

MLMR-COS includes five categories of ECFB, TR, SD, Normal and MH, each consisting of 193,251,83,280,82 groups, with 18 images in each group, and a total of 16002 images. Among them, 20, 20,10,20 and 10 groups are respectively taken as the test set. We take 10,10,5,10, and 5 groups of the remaining graphs in each category as the validation set, and the rest as the training set. The training set consists of 13,842 images.

**4.1.2. Evaluation metrics**

We use two commonly used metrics for evaluating the effects of image co-segmentation: Precision and Jaccard index. Precision is the percentage of correctly classified pixels in both background and foreground, which can be defined as

$$precision = \frac{|Segmentation \cap Ground\ truth|}{Segmentation}$$

Jaccard index (denoted by Jaccard in the following descriptions) is the overlapping rate of foreground between the segmentation result and the ground truth mask, which can be defined as

$$Jaccard = \frac{Segmentation \cap Ground\ truth}{Segmentation \cup Ground\ truth}$$

### 4.1.3. Implementation Details

For the preprocessing, all the images are normalized by the mean and the standard variation of the whole dataset. Because of limited computing resource, all images are resized to the resolution of 256 × 256 in advance. The co-segmentation results are resized back to the original image resolution for performance evaluation. We use the output of the image co-segmentation network proposed by Gong et al. [43] as the initial segmentation used in the first iteration, and input the initial segmentation to the RL co-segmentation network. In order to compare with the results of other state-of-the-art methods, we use MSRC and VOC2012 to train the network proposed by Gong et al, and use the results of testing MSRC and VOC2012 with the network of Gong et al as the training set of our RL-CoSeg method, and the results of testing the Internet dataset, ICoseg dataset and MLMR-COS are used as the testing set of our RL-CoSeg method. This leads to a different accuracy of the segmentation probability of the training set and the testing set. The segmentation quality of the training set is higher, and the segmentation quality of the test set is lower. It is equivalent to adding more noise to the testing set for the RL-CoSeg method, but it can still achieve a satisfactory improvement effect.

The learning method of RL-CoSeg utilizes on-line learning, each epoch learns a pair of related images, and each epoch contains 4 steps to iteratively train the initial segmentation probability map. The learning rate is initialized to 0.001, the learning rate drops to the original 0.9 every 5000 epochs, using the Adam algorithm for optimization with a minibatch size of 1. We choose A = {±0.05,±0.1,±0.2,±0.4} as the ideal action set in our model[8]. For experiments using Pascal VOC 2012 and MSRC dataset, the maximum epoch is set to 60,000（Due to the machine GPU memory limitation, each epoch only trains a pair of related images, which is equivalent to train on the complete dataset for 9 times）. For experiments using the MLMR-COS dataset，the maximum epoch is set to 40，000（It is equivalent to train on the complete dataset for 9 times）.

4.2. Comparisons on the internet dataset

The Internet dataset is composed of 3 classes (aircraft, car, and house) of thousands of downloaded Internet images. Following the methods for comparison, we evaluate our approach on its widely used subset, in which each class has 100 images.

The resultant performances on the Internet dataset from our method (denoted by Ours) as well as the compared methods are listed in Table 1. From the results shown in Table 1, we can conclude that our co-segmentation method based on RL can obtain the best co-segmentation results at present. Our improvement in segmentation results occurs on all three classes and in both precision and Jaccard index. Compared with the initial segmentation, which is the current best result (from Gong et al), our network can obtain better segmentation results on each class, and improve the average Precision and Jaccard by 0.6% and 1.4%, respectively. Since our method mainly optimizes the edges and details of the segmentation, which occupy a small proportion in the whole image, it can prove that our method can effectively optimize the initial segmentation for each type of image in the Internet dataset.

Table 1. The performance comparisons on Internet datasets. The best results are bolded.

| Method | Car | | Horse | | Airplane | | Average | |
|---|---|---|---|---|---|---|---|---|
| | Precision | Jaccard | Precision | Jaccard | Precision | Jaccard | Precision | Jaccard |
| Quan et al. | 88.5 | 66.8 | 88.3 | 58.1 | 92.6 | 56.3 | 89.8 | 60.4 |
| Han et al. | 89.0 | 68.3 | 90.1 | 61.3 | 91.2 | 58.7 | 90.1 | 62.8 |
| Yuan et al. | 90.4 | 72.0 | 90.2 | 65.0 | 91.0 | 66.0 | 90.5 | 67.7 |
| Li et al. | 93.9 | 82.8 | 92.4 | 69.4 | 94.1 | 65.4 | 93.5 | 72.5 |
| Gong et al. | 95.9 | 89.4 | 92.6 | 71.4 | 96.3 | 80.5 | 94.9 | 80.4 |
| Ours | **96.4** | **90.3** | **93.3** | **73.6** | **96.9** | **81.6** | **95.5** | **81.8** |

Figure 4 shows some segmentation results of our method. Each set of images includes the original image, the original segmentation results, the segmentation improved by deep RL method and the detail enlargement. Enlarged views show the details and differences more clearly. Compared with the original segmentation, the edges and details are significantly improved. The edges of the foreground target object are finer, foreground /background pixels are more accurately classified, there are fewer background pixels classified as foreground objects, over-segmentation and under-segmentation are improved, edges are less jagged and smoother.

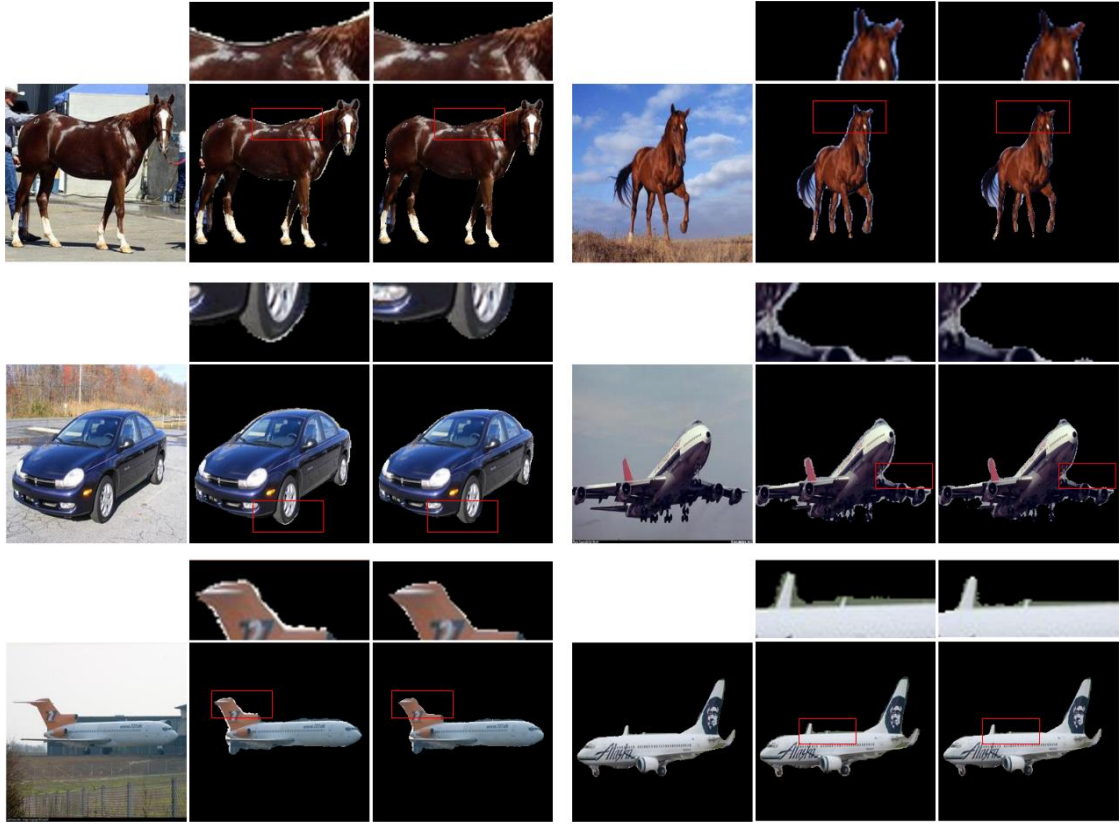

Figure 4. Segmentation results of Internet datasets. The figure contains six groups of test results. The first image in each group is the original image, the second image is the original segmentation result, and the third image is the result of the deep reinforcement learning co-segmentation network. The image above each group is a segmentation detail map.

4.3. Comparisons on the iCoseg dataset

Table 2. The comparisons of Jaccard index on iCoseg-subset.

| Class | Faktor and Irani | Jerripothula et al | Li et al | Gong et al | Ours |
|---|---|---|---|---|---|
| Bear2 | 72.0 | 67.5 | **88.7** | 86.2 | 86.8 |
| Brownbear | 92.0 | 72.5 | 91.5 | 92.7 | **93.2** |
| Cheetah | 67.0 | 78.0 | 71.5 | 91.1 | **91.5** |
| Elephant | 67.0 | 79.9 | 85.1 | 86.5 | **87.0** |
| Helicopter | 82.0 | 80.0 | 73.1 | 81.2 | **82.9** |
| Hotballoon | 88.0 | 80.2 | 91.1 | 94.5 | **94.7** |
| Panda1 | 70.0 | 72.2 | 87.5 | 92.9 | **93.2** |
| Panda2 | 55.0 | 61.4 | 84.7 | 88.9 | **89.1** |
| Average | 74.1 | 74.0 | 84.2 | 89.3 | **89.8** |

The entire iCoseg datasets contain 643 images divided into 38 object groups. In previous works, a subset of iCoseg (denoted by iCoseg-subset), involving 8 classes, was considered in Li et al' work based on the Siamese network [47]. In order to compare these methods with ours, we evaluate the proposed approach on not only the entire iCoseg but also the subset of it.

We test our RL-CoSeg method on the entire iCoseg and iCoseg-subset with 8 classes, and the results are shown in Table 2-3. We compare our performance with previous best ones from the methods based on deep learning and previous best ones from traditional methods. We can conclude that our method has reached the state-of-the-art level. In Table 2, for the Jaccard index, our method improves each class of the initial segmentation, and achieves the current best results in all classes except the bear class. On the entire iCoseg dataset, it brings the results very close to the currently best one, only -0.2% in Precision, and it reaches third-best Jaccard index.

In the entire iCoseg dataset, our method improves the results of Gong et al. Since the disparity between the segmentation method we used as the initial segmentation and the best method, there is a challenge for our method to surpass the best method. Our method uses only 2250 training examples. This amount is much less than that used in the work corresponding to the currently best result [26]. It is worth noting that for our RL co-segmentation method, there is a difference between the accuracy of the training dataset and the testing dataset. It is equivalent to adding noise to the testing dataset (detailed in section 4.1.3), which makes our segmentation task more difficult. Therefore, our method is expected to obtain finer results when the accuracy of the testing dataset is consistent with the training dataset.

Figure 5 shows some results in the iCoseg dataset segmented by our method. The foreground edge of the segmentation obtained by our method is more refined, contains fewer extra background pixels. In addition, our method also preserves more image details and adapts to the changes on the size, the pose, and the number of interested objects. As shown in the first set of images in Figure 5, our method segmented finer vertices and contours of the image.

Table3. The performance comparisons on iCoseg-entire

| Method | Precision | Jaccard |
|---|---|---|
| Faktor and Irani | 92.8 | 0.73 |
| Ma et al. | - | 0.79 |
| Han et al. | **94.4** | 0.78 |
| Yuan et al. | **94.4** | **0.82** |
| Gong et al | 94.1 | 0.77 |
| Ours | 94.2 | 0.78 |

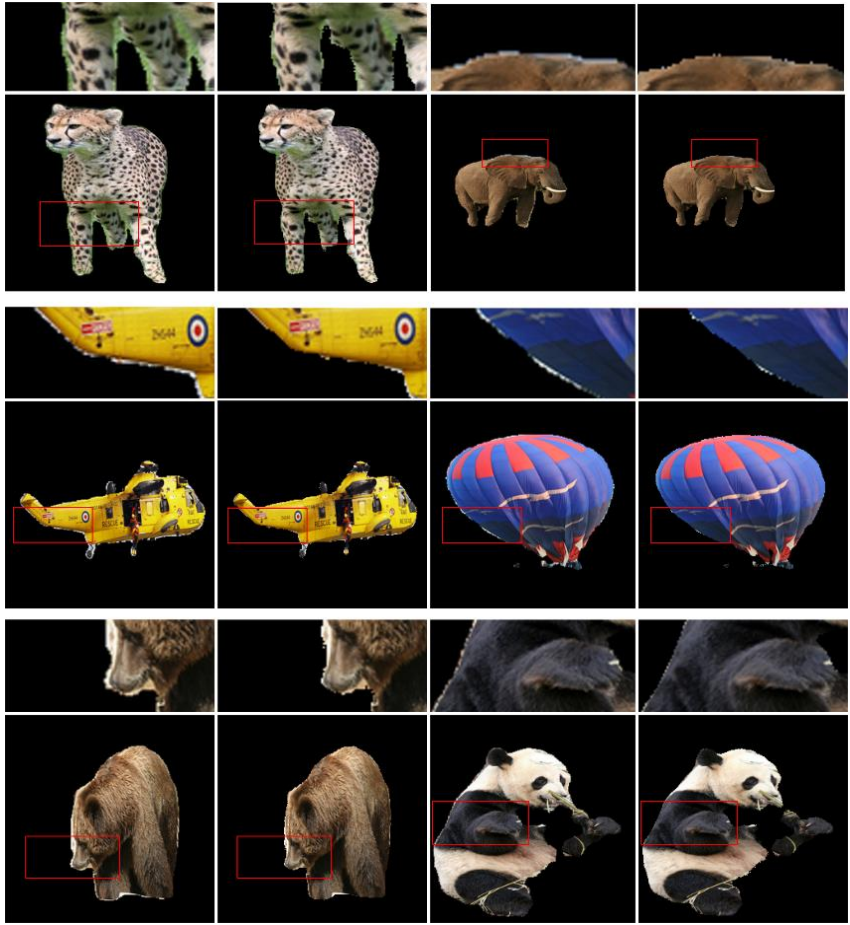

Figure 5. Segmentation results of the Icoseg dataset. The figure contains six groups for comparison. The first image of each group is the initial input of the RL co-segmentation method, and the second image is the result of the deep RL co-segmentation method.

4.4. Test on the MLMR-COS dataset

Table 4. Comparisons on the MLMR-COS dataset

|  | Normal | ECFB | MH | TP | SD | Mean |
| --- | --- | --- | --- | --- | --- | --- |
| Gong et al | 95.3 | 97.2 | 95.3 | 97.5 | 97.4 | 96.5 |
| Ours | **97.5** | **97.9** | **97.0** | **98.6** | **98.3** | **97.9** |

The initial segmentation of the Internet datasets and Icoseg datasets are relatively coarse. It is easier to improve coarse segmentation than fine segmentation. In order to verify the segmentation effect of our method on the fine initial segmentation, we use the MLMR-COS dataset as the testing dataset. MLMR-COS has a simple background generated by a single color. Because the foreground objects have transparent or white areas, it is easy to confuse the foreground and the background.

We use the segmentation results of Gong et al based on deep learning as the initial segmentation. Table 4 shows the comparison of our test results on the MLMR-COS dataset with the deep learning segmentation method of Gong et al. We can see that although the segmentation results of Gong et al for comparison are already fine, the space to improve the segmentation results is very small. It is difficult to improve the high-precision segmentation results, but our method can improve the Jaccard of each class of testing dataset. Our method improves on Normal and MH class by 2.2% and 2.3%, respectively. And the Jaccard of all classes is higher than 97%, especially TP and SD exceed 98%, which reaches an accurate level.

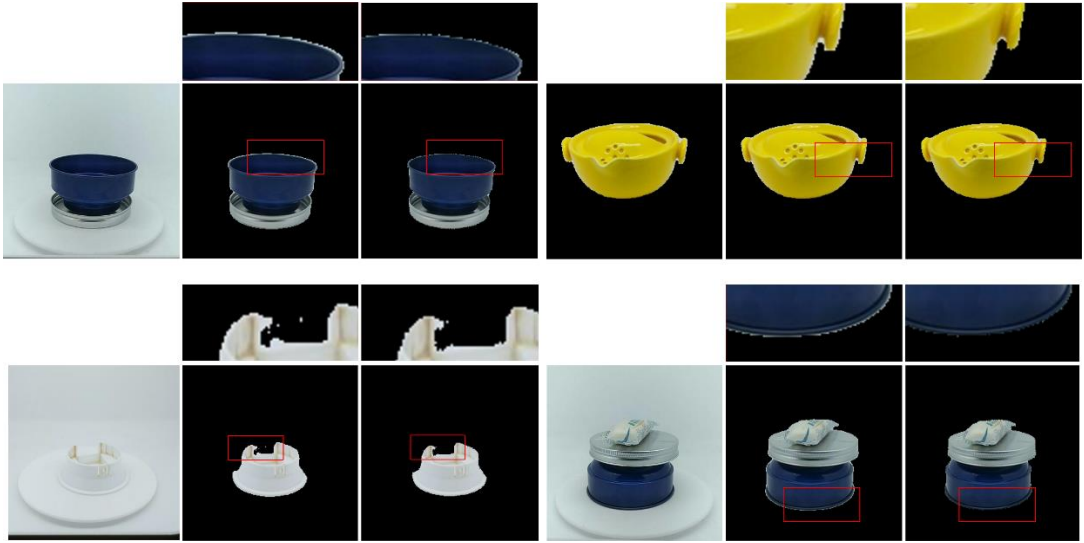

Figure 6. Segmentation of the MLMR-COS dataset. The figure contains four groups of test results. The first image in each group is the original image, the second image is the original segmentation result, and the third image is the result of the deep reinforcement learning co-segmentation network. The image above each group is a segmentation detail map.

Figure 6 shows the qualitative comparison between the initial segmentation obtained by method of Gong et al and the segmentation obtained by our method. Among them, the small window around the edge of the foreground is enlarged to show the difference more clearly. It can be observed that the edges of the segmentation obtained by our method are more detailed, removing the obvious white edges in the original segmentation and retaining more details. This proves that our method performs better in capturing edge details.

4.5 Ablation study

Table 5. The comparisons on Internet.

| Method | Precision | Jaccard |
| --- | --- | --- |
| -Co-Segmention | 96.6 | 89.1 |
| -Self-Attention | 96.9 | 89.5 |
| Full network | **97.1** | **89.8** |

In order to justify the design of our RL image co-segmentation method is reasonable, we make the following changes to our segmentation network and compare the performance of these modified versions:

1）-Self-Attention：Only use co-attention block based on Siamese network structure for RL image co-segmentation without Self-Attention module.

2）-Co-Segmention：Remove the Self-Attention module and the co-segmentation structure. The encoder structure and decoder structure are consistent with the original network.

We compare these two changes with our complete network. The performance comparison of the three networks is shown in Table 5, which illuminates that:

1）Compared with the complete network, the improved Self-Attention block for the RL method provides more information, and the long-distance dependence of the image helps to improve the segmentation of the network.

2）Both Precision and Jaccard are reduced without the co-segmentation structure, which proves that the important performance of deep RL co-segmentation algorithm. For our dataset, the RL co-segmentation structure achieves finer results.

3）The RL image co-segmentation algorithm can meet the needs of the co-segmentation task and obtain fine segmentation.

## 5. Conclusions

This paper proposes an automatic reinforcement learning image co-segmentation algorithm. For the problem of co-segmentation, this paper designs an A3C algorithm that utilizes the correlation between images. In order to obtain the co-attention of two related images in co-segmentation, we construct a Siamese RL co-segmentation network, which has policy branches and value branches to output the action and value for each image. To get finer segmentation and obtain more image feature to supplement the deleted user interaction, we improve self-attention for the RL method and embed it in the network structure. Our network is trained on the MSRC and PASCAL VOC 2012 datasets, and tested on the Internet, iCoseg and MLMR-COS datasets. Ablation study verifies the effectiveness of our self-attention and the effectiveness of the Siamese deep RL co-segmentation algorithm. Compared with the previous methods, our RL-CoSeg method has achieved the best performance on Internet datasets and the most advanced performance on iCoseg-subset, it improve the initial segmentation on iCoseg-entire, and it achieve high accuracy in MLMR-COS dataset. RL-CoSeg method can obtain a finer co-segmentation result and keep more details. Especially the segmentation edge has been greatly improved. To further improve the performance of this method, increasing the amount of training dataset and making the precision of the training dataset and the testing dataset more consistent are possible solutions, which we will discuss in future work.